\documentclass[]{fairmeta}

\usepackage{graphicx}
\usepackage{amsmath}
\usepackage{amssymb}
\usepackage{booktabs} 
\usepackage{algorithm}
\usepackage{algpseudocode}
\usepackage{multirow}
\newcommand*{\Scale}[2][4]{\scalebox{#1}{$#2$}}%

\newcommand{\minisection}[1]{\vspace{1mm}\noindent{\textbf{#1}.}}

\title{SiamMM: A Mixture Model Perspective on Deep Unsupervised Learning}

\author[1]{Xiaodong Wang}
\author[1]{Jing Huang}
\author[1]{Kevin J Liang}

\affiliation[1]{Meta AI}

\abstract{
   Recent studies have demonstrated the effectiveness of clustering-based approaches for self-supervised and unsupervised learning. However, the application of clustering is often heuristic, and the optimal methodology remains unclear. In this work, we establish connections between these unsupervised clustering methods and classical mixture models from statistics. Through this framework, we demonstrate significant enhancements to these clustering methods, leading to the development of a novel model named SiamMM. Our method attains state-of-the-art performance across various self-supervised learning benchmarks. Inspection of the learned clusters reveals a strong resemblance to unseen ground truth labels, uncovering potential instances of mislabeling. 
}

\date{October 24, 2024}
\correspondence{First Author at \email{xiaodongwang@meta.com}}

\begin{document}

\maketitle

\section{Introduction}
\label{sec:intro}

Recent self-supervised and unsupervised learning methods have demonstrated the ability to learn representations that are surprisingly competitive with fully supervised learning~\cite{SwAV, SimSiam, PIRL, MoCov2, Barlow}.
Rather than relying on human annotations, which can be costly or biased, recent approaches have used Siamese network architectures with implicit labels: augmentations of the same image instance are considered positive samples, while any other image is considered to be negative.
This instance-based approach has proven successful, but ignores the semantic similarities present in the image data.
For example, ImageNet~\cite{deng2009imagenet}, a commonly used dataset for unsupervised learning, is actually a supervised dataset purposefully collected from a hierarchical set of labels~\cite{miller1995wordnet}, and even non-curated image data contain examples with similar characteristics~\cite{pope2021intrinsic}.
Disregarding this structure can lead to similar images repelling each other in the embedding space, hampering the self-supervised learning process.
To address this, several recent works have incorporated clustering strategies~\cite{Deepcluster, SwAV, PCL, Propos}.
Although learned clusters have been shown to lead to strong performance in various downstream tasks, several design choices of these past clustering methods, often based somewhat heuristically on $K$-means~\cite{lloyd1982least}, remain unclear.

To enhance our understanding of clustering methods, we leverage a principled statistical foundation and conceptualize clustering-based self-supervised learning as a mixture model (MM). Mixture models characterize data as a combination of component distributions, with each component representing a distinct sub-population within the overall distribution. Given the semantic structure inherent in the image data, we posit that MMs serve as a natural fit for unsupervised representation learning.
Depending on whether embeddings are normalized, we cast clustering in representation learning as either a Gaussian mixture model (GMM) or von Mises-Fisher mixture model (vMFMM). 

\begin{figure}[t]
\label{fig.loss}
\begin{center}
\centerline{\includegraphics[scale=0.35]{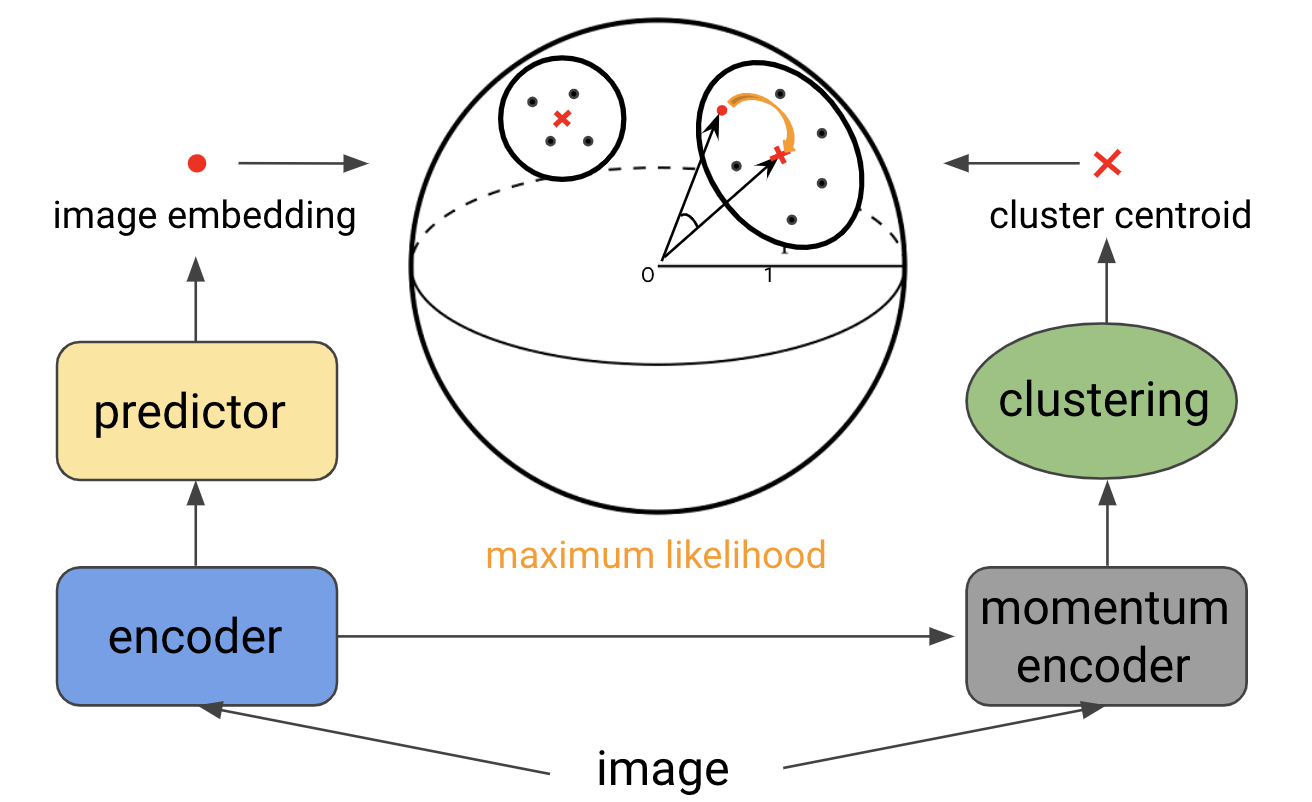}}
\caption{SiamMM model architecture. Assuming all the embeddings are $L^2$ normalized, we cast clustering in representation learning as a von Mises-Fisher mixture model (vMFMM). The optimization objective tends to minimize the distance between an embedding and its clustering centroid (or nearest centroids) without negative samples. }
\end{center}
\end{figure}

Examining previous methodologies through this lens allows us to discern the underlying techniques and perceive improvements within more generalized mixture models. For example, $K$-means, often employed in cluster-based self-supervised learning through contrastive loss with a large number of clusters, can be viewed as an empirical execution within this broader framework.
By relaxing some of these constraints or drawing analogies to common Expectation Maximization (EM) procedures for learning mixture models, we gain insights into more effective ways of utilizing clustering in self-supervised learning. The mixture model formulation also suggests alternative procedures, such as consistent clustering, cluster variance, and soft assignment, as opposed to the prevalent methods that involve re-initialized clustering, equal spherical clusters, and hard assignment.

Another fundamental question in the clustering or mixture model is determining the number of clusters, which is often picked by heuristic intuition, cross-validation, or information criterion~\cite{inform}. In the context of deep representation learning, previous unsupervised learning methods tend to overcluster the data population to achieve high model performance~\cite{PCL}, which results in quite a lot of clusters where some of them only capture a single point inside. We argue that this phenomenon contrasts with the original motivation of clustering semantically similar subpopulations. 
In order to disentangle the uncertainty in determining the number of cluster components, we seek a non-parametric solution to merge clusters throughout the training process. Although a simpler model is obtained, the performance achieves state-of-the-art performance on a series of self-supervised learning benchmarks.

We suggest changes to clustering in representation learning to reflect the insights of a mixture model perspective. 
Because of this inspiration, we name our method \textbf{SiamMM}.
We conducted a number of self-supervised learning experiments on ImageNet~\cite{deng2009imagenet} with ResNet architectures~\cite{he2016deep}, demonstrating that our perspective of the mixture model indeed leads to empirical improvements. Our main contributions are summarized as follows: 

\noindent1. We introduce a new representation learning approach, SiamMM, which interprets clustering as a statistical mixture model and applies its insights in a self-supervised manner. Our MLE-based loss function is more adaptive and improves the accuracy of Siamese networks without the need to sample negative samples. \\2. We propose a novel scheme that dynamically reduces the number of clusters during pretraining. This reduction not only cuts down on training time but also facilitates interpretation. The learned clusters exhibit a strong resemblance to the unseen ground truth labels for ImageNet. \\3. SiamMM achieves advancements over previous methods across various Self-Supervised Learning (SSL) benchmarks. We further explore the impact of negative samples in clustering, consistent centroid updates, cluster concentration estimation, and soft cluster assignment through an extensive ablation study. We hope our valuable insights can propel the field of unsupervised learning forward.

\section{Related work}

\minisection{Self-supervised contrastive learning} Contrastive learning \cite{contrast} approaches to self-supervised learning in computer vision aim to learn an embedding space where views or augmented crops of an image have more similar embeddings than those of different images. Many contrastive methods rely on mechanisms of sampling a large number of negative samples \cite{SimCLR, MoCo, Tian2, Wu2018} to prevent the model from learning trivial solutions with unlabeled data. For example, SimCLR \cite{SimCLR} requires a large batch size to have enough negative samples; MoCo \cite{MoCo} maintains negative samples in a memory bank \cite{Wu2018} and consistently updates the embeddings via a separate momentum encoder.

Negative samples were initially thought to prevent representation collapse: without something to counteract the gradients pulling embeddings together, it was thought that self-supervised methods would lead to the trivial solution of all inputs mapping to a constant embedding. However, a recent series of methods \cite{BYOL, SimSiam, Barlow, vicreg} have shown it possible to directly learn invariant features from distorted versions of an image without any negative samples. BYOL \cite{BYOL} and SimSiam \cite{SimSiam} introduce asymmetry in the network and learn updates using one distorted version and a stop gradient operation on the other version. Barlow Twins \cite{Barlow} and VICReg \cite{vicreg} add regularization in the cross-correlation matrix of embeddings of positive samples to reduce over-fitting of redundant parameters.

\minisection{Deep clustering}
Clustering-based methods for self-supervised learning have seen success in self-supervised learning \cite{SeLa, Caron19, Deepcluster, SwAV, PCL}. Though negative samples are not directly used, pseudo-labels or cluster centroids play a role in making a contrastive loss. DeepCluster \cite{Deepcluster} and SeLa \cite{SeLa} cluster representation embeddings from previous iterations and train a classifier based on the cluster index as the pseudo-label. SwAV \cite{SwAV}, ProPos\cite{Propos}, and PCL \cite{PCL} incorporate negative cluster centroids in a contrastive loss, though SwAV and ProPos pulls the centroid embeddings of two distorted versions of an image closer, while PCL uses one distorted version to predict its cluster centroid computed from another version. Another difference among these clustering-based methods is that DeepCluster and PCL adopt $K$-means in cluster analysis, while SeLa and SwAV perform online clustering via Sinkhorn-Knopp \cite{Sinkhorn}.

\minisection{Mixture Models}
Mixture models are common statistical models that have been in use for more than a century, with Gaussian mixture models (GMMs) a popular choice~\cite{mclachlan2000finite, day1969estimating}.
These models are particularly adept at modeling data with sub-populations, which tend to form clusters within the overall distribution. As an unsupervised learning approach, GMM and its variant can lie well in Deep Neural Network for density estimation~\cite{NIPS2014_8c3039bd}, clustering~\cite{9010011}, sampling~\cite{Gepperth2021ImageMW}, and anomaly detection~\cite{zong2018deep}, and has been successfully applied in multiple fields of computer vision, such as image processing~\cite{restoration,compression}, segmentation~\cite{GUPTA1998315}, generation~\cite{Gepperth2022ANP}, and face verification~\cite{face_veri}. The interpretation of clustering in self-supervised representation learning as a mixture model 
are relatively limited in the literature.

\section{Background}
\subsection{Siamese Networks}
Representation learning methods seek to learn a mapping from an image $x \in \mathcal X$ to a $d$-dimensional embedding $v \in \mathcal V$, such that the embedding space $\mathcal V$ captures useful information and structure applicable to downstream tasks.
This mapping is often parameterized as a deep neural network $\mathcal E(\cdot)$, with $v=\mathcal{E}(x)$. 

Recent works often adopt a Siamese framework, seeking to maximize the embedding similarity between two augmentations $x_1$ and $x_2$ of a single image $x$. 
An encoder followed by a small multilayer perceptron (MLP) predictor $g$ generates embedding $v_1 = g(\mathcal E(x_1))$ on one branch, while on the other branch, momentum encoder $\mathcal E_m(\cdot)$ generates a second embedding $v_2^m = \mathcal E_m(x_2)$.
The loss is defined as 
\begin{equation}
\label{eq:simsiam_loss}
\mathcal L(v_1, v_2^m) = D(v_1,\mathrm{stopgrad}(v_2^m))
\end{equation}
where $D$ is a distance metric defined in the embedding space and $\mathrm{stopgrad}(\cdot)$ is a stop-gradient operation applied to the second view's embedding and $D$.

\subsection{Clustering} 
As posed in (\ref{eq:simsiam_loss}), each instance is considered to be an independent entity, with the loss only enforcing similarity between augmented views for a particular example.
For methods with contrastive losses, embeddings of all other samples are additionally pushed away.
Since this may not be ideal if there exists many subpopulations within the data distribution, clustering methods allow grouping samples by semantic similarity.
Given $N$ samples $ x_i \in \mathcal{X}, 1 \leq i \leq N$ and their corresponding embedding vectors $v_i = \mathcal{E}(x_i) \in \mathcal{V}$, a clustering method will produce a cluster assignment $\mathcal{A}(x_i) \in \{1, 2, ..., K\}$, and $\mathrm{C}_k := \{v_i \in \mathcal{V} | \mathcal{A}(x_i)=k, 1 \leq i \leq N\},  1 \leq k \leq K$ indicates the embedding point cloud of the $k$-th cluster.

\subsection{Mixture Models}
The Gaussian Mixture Model (GMM) is a statistical method commonly used in modeling probabilistic clustering and is widely applied in data mining and machine learning~\cite{murphy2013machine}. 
Given the cluster $\mathrm{C}_k$, assume that the image embeddings $v \in \mathrm{C}_k$ follow a Gaussian distribution centered at $\mu_k$ with variance $\Sigma_k$. The mixture distribution of $K$ Gaussian clusters is given by the summation:
\begin{equation}
\label{eq:GMM}
\begin{split}
  f(v) = \sum_{k=1}^K \pi_k f_k(v|\mu_k , \Sigma_k)
 \end{split}
\end{equation}
where $f_k$ is the density function of a $d$-dimensional Gaussian distribution of the $k$-th cluster $\mathrm{C}_k$; $\pi_k$ is the prior probability of $\mathrm{C}_k$.

Due to the common practice of $L^2$ normalization, the embedding $v$ lies on a hypersphere and thus can more appropriately be modeled with a von Mises-Fisher mixture model (vMFMM). Suppose that $v$ is a $d-$dimensional unit random variable, i.e. $v \in \mathcal{R}^d, ||v||=1$. The likelihood function that $v$ is generated from $f_k(.)$ is
\begin{equation}
\label{eq:f_k}
\begin{split}
  f_k(v | \mu_k , \kappa_k) = c_d(\kappa_k) \exp{(\kappa_k \mu_k^T v)}
 \end{split}
\end{equation}
where $\mu$ is a mean direction with $||\mu||=1$, $\kappa$ is the concentration, and normalization term $c_d(\kappa)$ is given by
\begin{equation}
\label{eq:c_d}
\begin{split}
  c_d(\kappa) = \frac{\kappa^{d/2-1}}{(2\pi)^{d/2}I_{d/2-1}(\kappa)}
 \end{split}
\end{equation}
where $I_d(.)$ is the modified Bessel function of the first kind and order $d$.


\minisection{Expectation Maximization (EM) algorithm} The parameters of mixture models are commonly estimated with the EM algorithm, by maximizing the log-likelihood function. In the E-step, the probability that $v$ is assigned to the cluster index $k$ is calculated, denoted as $p(k|v)$. 
Let the unnormalized mean vector of cluster $k$ be defined as
\begin{equation}
\label{eq:r_k}
\begin{split}
  r_k = \frac{\sum_{i=1}^N v_i p(k|v_i)}{\sum_{i=1}^N p(k|v_i)}
 \end{split}
\end{equation}
Then, $\mu$ and $\kappa$ can be approximated by Maximum Likelihood Estimation (MLE) \cite{vMF}:
\begin{align}
  \hat{\mu_k} &= \frac{r_k}{||r_k||} \label{eq:mu} \\ 
  \hat{\kappa_k} &= \frac{||r_k||d - ||r_k||^3}{1-||r_k||^2} \label{eq:kappa}
\end{align}
for $k=1,2,...,K$.

\section{SiamMM for Clustering Representations}


We propose viewing clustering in unsupervised representation learning through the lens of mixture models.
Under such a paradigm, each cluster can be viewed as a mixture component, representing a subpopulation of the data distribution. 
Though inspired by mixture models, we do not directly model the learned embeddings as a standard mixture model, as there are several key differences of the deep representation setting.
Firstly, the goal of representation learning is to learn the encoder $\mathcal E$, whose parameter updates must occur in conjunction with the EM learning of the mixture model parameters.
Learning the mapping of $\mathcal X \mapsto \mathcal V$ also means that the deep feature distribution modeled by the MM is constantly evolving, in contrast to the fixed data features commonly modeled by mixture models in statistics.
Instead, we primarily use mixture models to motivate changes in the common clustering methodology.


To adapt mixture models to deep representation learning, we adopt a two-tier EM algorithm to model cluster distributions and learn representation embeddings simultaneously.
In the E step, the assignment function is updated based on the current centroids, and so is the mean vector of ($\ref{eq:r_k}$).
This is followed by an M step to estimate the  mixture model parameters $\mu$ and $\kappa$ via MLE (\ref{eq:mu}, \ref{eq:kappa}), given the current parameters of $\mathcal{E}$.
These E and M steps are iterated, updating the cluster model.
We then perform an outer loop M step, updating $\mathcal{E}$ to maximize the likelihood of the data.
We provide a deeper dive into changes to individual components below.


\subsection{MLE: Non-negative Soft-assignment Loss}
\label{sec:neg_samp}
Previous methods have posed a variety of loss objectives for clustering, including pseudo cluster label prediction~\cite{Deepcluster}, swapped assignment prediction~\cite{SwAV}, or directly replacing one of the views~\cite{PCL} or both views~\cite{Propos} with cluster centroids in the contrastive loss. 
Several of these methods utilize negative centroids as an additional repelling force between clusters.

Additionally, many previous clustering approaches assume a hard-assignment strategy, where the assignment probability $\pi_k$ is implicitly treated as a binary 0-1 variable given by
\begin{equation}
\label{eq:A_1}
\pi_k(v) =
    \begin{cases}
      1 & \text{if $v \in \mathrm{C}_k$}\\
      0 & \text{otherwise}
    \end{cases}       
\end{equation}
In the case that $v \in \mathrm{C}_k$, the density function of mixture models (\ref{eq:GMM}) reduces to a single distribution $f_k$.

While hard assignment simplifies clustering, it is also an assumption that limits the model expressivity. Alternatively, we can generalize and allow the mixture model to have \textit{soft} assignment. 
Within the context of representation learning, this allows a single sample to be driven towards multiple prototypes, with varying probability. 
In practice, we limit soft assignment to a set of $H$ nearest centroids around $v$. Suppose that the set of nearest centroids is $\mathcal{M}$ with size $|\mathcal{M}|=H$, so $\pi_k(v) = 0$ for $k \notin \mathcal{M}$. 

In mixture models, the M step of the EM algorithm typically estimates the model parameters with maximum likelihood estimation (MLE). The negative log-likelihood can be expressed as the following weighted summation:
\begin{equation}
\label{eq:soft}
\begin{split}
   \mathcal{L}^H(v|\hat{\kappa}_{\mathcal{A}(x)}, \hat{\mu}_{\mathcal{A}(x)}) = -\log\sum_{k \in \mathcal{M}} \pi_k \exp(\hat{\kappa}_k \hat{\mu}_k^T v)
 \end{split}
\end{equation}
where $\hat{\kappa}$ and $\hat{\mu}$ are now fixed estimators and carry no gradient operation; $\pi_k$ is the weight scale for the $k$-th assignment. We define $\pi_k$ by considering the relative similarity score between $v$ and its centroids by
\begin{equation}
\label{eq:w_h}
\begin{split}
   \pi_k(v) = \frac{\alpha_k \exp(\mu_k^T v/\tau)}{\sum_{l}^H \alpha_l \exp(\mu_l^T v/\tau)}
 \end{split}
\end{equation}
where $\alpha_k$ is a measure of the size of cluster; $\tau>0$ is a tuning parameter; for a large temperature $\tau$, the weight tends to be evenly distributed over $H$; for a small $\tau$, the weight lies more on the closest centroids.
Note that in the special case where $H=1$, this soft assignment reduces to the hard assignment of (\ref{eq:A_1}).



Notably, MLE does not rely on any negative sampling; instead, one only needs to maximize the similarity between embeddings and their nearest clusters.
By removing any notion of negatives from our objective functions, we can improve computational efficiency by avoiding computing additional negatives terms in the loss.
Note that this corroborates recent works finding that negative sampling is unnecessary for non-clustering self-supervised methods~\cite{BYOL, SimSiam, Barlow, vicreg}.

\subsection{Cluster Merging}
\label{sec:merge}
One of the prevalent challenges in many mixture models or clustering algorithms is the determination of the number of clusters, often unknown \textit{a priori} and requiring estimation through methods such as cross-validation or analytic criteria. In the realm of self-supervised learning, this challenge is compounded by the presence of embedding shift: as the encoder is trained, the distribution and clustering behavior of embeddings change over time. Although employing a large number of clusters (K) tends to yield strong empirical performance, it can compromise the interpretability of clustering results. For instance, advocating for $100k$ clusters, as suggested in~\cite{PCL}, is a hundred times larger than the number of classes in ImageNet. Furthermore, a fixed high value of K increases computational costs, underscoring the importance of minimizing K when feasible.

Instead of manually specifying the number of clusters, we propose a merging strategy that dynamically adapt the parameter K to the evolving embedding space. In the initial stages, when the representations lack substantial meaning, it is advantageous to commence with a larger number of clusters. Subsequently, at each iteration, we refine the clustering by \textit{merging} the closest clusters by a predefined threshold $\zeta$. Suppose the pair-wise distance between cluster index $i$ and $j$ is $\mathcal{Z}_{ij} := ||\mu_i - \mu_j ||_2$, then the two cluster are merged if
\begin{equation}\label{eq:merge}
    \frac{\mathcal{Z}_{ij} - E\{\mathcal{Z}_{ij}\}}{SE\{\mathcal{Z}_{ij}\}} < \zeta
\end{equation}
where $E\{\mathcal{.}\}$ and $SE\{\mathcal{.}\}$ are the expectation and standard deviation of the pair-wise distance, respectively. This approach, characterized by the initial use of a large number of clusters followed by a gradual reduction to a smaller set of key centroids, resonates with the principles of the Lottery Ticket Hypothesis~\cite{frankle2019lottery}: by initially overseeding clusters, there are increased chances of identifying the ``correct'' centroids, thereby enhancing clustering performance. Such an approach mitigates the computational costs associated with a fixed large number of clusters, ultimately converging to a number on par with the number of classes in ImageNet without sacrificing accuracy.

\subsection{Consistent Centroids Update}
\label{sec:consistent}
There is a well-known tendency of $K$-means to fall into poor local minima, and the final result is often sensitive to centroid initialization~\cite{pena1999empirical}.
In practice, it is common to run many re-initializations and pick the best result~\cite{jain1999data, PCL}.

The practice of re-initializing clustering after each epoch results in several negative side effects in self-supervised learning.
Firstly, these re-initializations mean that the cluster centroids are constantly changing, beyond just the embeddings being updated by backpropagation; the very clustering structure, which is meant to represent learned semantic structure in the data, is inconsistent, leading to a learning signal full of thrash.
In contrast, we observe that the mixture model formulation suggests that the mixture distributions should be learned consistently.
Rather than re-initializing from scratch multiple times each epoch, centroids should be initialized exactly once, at the start of training; subsequent clustering per epoch should be initialized with the previous epoch's centroids.
By doing so, we provide the algorithm with more consistent cluster targets, and removing the need for multiple restarts significantly speeds up training. The difference between one round vs. five random initializations~\cite{PCL} per epoch is an approximately 25$\%$ increase in training time, even with an efficient implementation like Faiss~\cite{faiss}.

\subsection{Non-uniform Cluster Size}
$K$-means is well known for producing clusters of roughly equal size,
as every point in $\mathcal V$ is assigned to the nearest centroid, without any notion of the learned per-cluster size.
This may not be a good assumption in certain applications.
For example, while the commonly used ImageNet-1K dataset~\cite{deng2009imagenet} has a relatively balanced number of samples per class, the classes themselves are not necessarily evenly distributed semantically (e.g., over $10\%$ of the categories are breeds of dogs), nor are the clusters one might recover when learning unsupervised structure.

In the context of the von Mises-Fisher mixture model, each component of the mixture is characterized not only by a mean $\mu_k$ (i.e., the centroid) but also by a concentration parameter per cluster $\kappa_k$. The ability to learn per-cluster variance allows for a more expressive representation of the embedding distribution. 

\subsection{Final Loss Function}
Clustering algorithms like $K$-means may fail to converge at the beginning of the training process, when the representations have yet to learn much meaning. 
Empirically, we found that adding an instance-wise loss can boost clustering efficiency and improve performance. This finding is consistent to the additive loss suggested by \cite{PCL,SLIC,Propos, CC}. 
We define the symmetrized instance-wise loss from (\ref{eq:simsiam_loss}) as
\begin{equation}
\label{eq:inst_loss}
\mathcal{L}_{inst} = \mathcal{L}(v_1, v_2^m) + \mathcal{L}(v_2, v_1^m)
\end{equation}
This instance-wise loss is combined with the cluster-wise loss, specifically the soft-assignment loss $\mathcal{L}^H$ in (\ref{eq:soft}). The final loss for SiamMM takes an additive form
\begin{equation}\label{eq:tol_loss}
\mathcal{L}_{final} = \mathcal{L}^H + \mathcal{L}_{inst}
\end{equation}
Importantly, neither of these terms relies on negative samples.


\section{Experiments}
\subsection{Implementation details}
We pre-train SiamMM on the ImageNet-1k dataset~\cite{deng2009imagenet}, with a standard ResNet-50~\cite{he2016deep} as encoder $\mathcal E$. The projection head is a 3-layer MLP, and the prediction head is a 2-layer MLP. Following \cite{mocov3}, all MLP layers have Batch Normalization~\cite{ioffe2015batch}, with 4096 dimensional hidden layers with ReLU activations, and the output layers are 256 dimensional without ReLU.

Our training procedure proceeds as follows: First, we update the cluster assignments using the centroids computed from the previous epoch, and estimate the mixture model parameters 
$\mu$ and $\kappa$ using equations (\ref{eq:mu}) and (\ref{eq:kappa}), respectively. Next, we backpropagate through the encoder $\mathcal E$ using the loss function in (\ref{eq:tol_loss}). Finally, we apply the cluster merging strategy described in Section~\ref{sec:merge} and update the cluster centroids accordingly.

We use Faiss \cite{faiss} for clustering and the nearest centroid search, with a default initial number of $K=100k$ for cluster merging. We find that setting the merging threshold $\zeta=-1.2$ in (\ref{eq:merge}), which corresponds to the $10$-th percentile of the normalized pairwise distance between cluster centroids, yields a promising merging curve. 
This threshold works well for large cluster numbers, though it may vary with data distribution and initial cluster count. We recommend that researchers adjust this parameter through methods like grid search to suit their specific datasets and clustering needs.
We set $\alpha=1$ and $\tau=0.02$ in (\ref{eq:w_h}) according to soft assignment when $H>1$; by default, $H=5$ unless otherwise stated. In the estimation of concentration parameter $\kappa$, we find it beneficial to apply principal component analysis (PCA) to reduce dimensional correlations in the features \cite{dim_collapse}. More details on this process can be found in Appendix~\ref{apx:pca}.

We adopt the same configuration of data augmentations as MoCo v3~\cite{mocov3}. We use the LARS optimizer~\cite{lars} by default with a weight decay of 1e-6, a momentum of 0.9, batch size 4096, and a base learning rate $lr_{base}=0.6$ and $0.5$ for 100-epoch and 200-epoch, respectively. The learning rate is adjusted for different batch size $lr = lr_{base} \times batch\_size / 256$ and follows a cosine decay schedule.

\subsection{Image classification benchmarks}
We evaluated the pre-trained ResNet-50 backbone with SiamMM for 100 epochs and 200 epochs on ImageNet. 

\minisection{ImageNet Linear Evaluation} We compare SiamMM's learned representations with recent state-of-art methods by training a linear classifier on top of a frozen ResNet-50 backbone. The validation set top-1 accuracies are reported in Table~\ref{table:imagenet_linear_eval}. We use an SGD optimizer with cosine learning rate with $lr_{base} = 0.1$, momentum of $0.9$, weight decay of $0.0$, and batch size of $1024$, for $90$ epochs. The 200-epoch pretraining results of BYOL~\cite{BYOL} and SwAV~\cite{SwAV} are from the implementation in \cite{SimSiam}, which reproduces the results with two 224$\times$224 crops of each image for a fair comparison. We observe that SiamMM outperforms the previous methods when considering a similar number of pre-training epochs. 
In particular, SiamMM outperforming other clustering methods such as SwAV~\cite{SwAV}, PCL~\cite{PCL}, and ProPos~\cite{Propos} demonstrate the value of our perspective of the mixture model.

\begin{table}[t]
\caption{\textbf{Linear classification on ImageNet.} We report top-1 accuracy (in \%) with a frozen pretrained ResNet-50. 
†: result reproduced by us.}
\label{table:imagenet_linear_eval}
\begin{center}
\begin{small}
\begin{sc}
\begin{tabular}{lcccc}
\toprule
Method                         & Epochs & Top-1  & Epochs & Top-1                 \\ \midrule
SimCLR\cite{SimCLR}                       & 100  & 64.6 & 200   & 66.8                   \\
MoCo v2\cite{MoCov2}                      & 100  & - & 200 & 67.5                        \\
PCL\cite{PCL}                             & 100  & - & 200   & 67.6                      \\
SwAV\cite{SwAV}                           & 100  & 66.5 & 200   & 69.1                   \\
SimSiam\cite{SimSiam}                     & 100  & 68.1 & 200   & 70.0                   \\
BYOL\cite{BYOL}                           & 100  & 66.5 & 200 & 70.6                     \\
All4One\cite{All4One}                     & 100  & 66.6 & 200 & -                        \\
NNCLR\cite{NNCLR}                         & 100  & 69.2 & 200 & 70.7                     \\
MoCo v3\cite{mocov3}                      & 100  & 68.9 & 200   & 71.0 †                 \\
ProPos\cite{Propos}                       & 100  & -    & 200   & 72.2                   \\
SNCLR\cite{SNCLR}                         & 100  & 69.6 & 200   & 72.4                   \\
LEWEL\cite{Lewel}                         & 100  & 71.9 & 200   &   72.8                 \\
SiamMM (ours)                             & 100  & \textbf{71.9} & 200 & \textbf{73.2}   \\
\bottomrule
\end{tabular}
\end{sc}
\end{small}
\end{center}
\end{table}

\minisection{Semi-supervised Learning}
We also evaluate the performance of the pretrained backbone by fine-tuning it with a linear classifier using 1\% and 10\% of the labels. We use the split of \cite{SimCLR} to select a subset of ImageNet training data with labels and report the top-5 accuracy on the ImageNet validation set in Table~\ref{semi-sup}. Our method is competitive with the state-of-the-art even compared with the previous methods pre-trained on a larger number of epochs. 

\begin{table}[t]
\caption{\textbf{Other downstream evaluation tasks.} (left) semi-supervised evaluation for ResNet-50 backbone fine-tuned on 1\% and 10\% ImageNet data; the top-5 accuracy are reported (in \%). (right) transfer learning results for other downstream tasks, Places205 \cite{zhou2014learning} scene classification with top-1 accuracy, and VOC07 \cite{everingham2009pascal} multi-label image classification with mAP.}
\label{semi-sup}
\begin{center}
\begin{small}
\begin{sc}
\begin{tabular}{lccccc}
\toprule
\multirow{2}{*}{Method} & \multirow{2}{*}{Epochs} & \multicolumn{2}{c}{ImageNet}                                    & \multirow{2}{*}{Places205}     & \multirow{2}{*}{VOC07}         \\
                        &                         & 1\%                            & 10\%                           &                                &                                \\  \midrule
SimCLR\cite{SimCLR}                                      & 200                     & 56.5                           & 82.7                           & -                              & -                              \\
MoCo v2\cite{MoCov2}                                     & 200                     & 66.3                           & 84.4                           & -                              & -                              \\
PCL\cite{PCL}                                      & 200                     & 73.9                           & 85.0                           & 50.3                           & 85.4                           \\
NNCLR\cite{NNCLR}      & 400 & 79.2 & 88.6 & - & - \\
SNCLR\cite{SNCLR}      & 400 & 80.1 & 89.1 & - & - \\
PIRL\cite{PIRL}                                & 800                     & 57.2                           & 83.8                           & 49.8                           & 81.1                           \\
SimCLR\cite{SimCLR}                                      & 800                     & 75.5                           & 87.8                           & 53.3                           & 86.4                           \\
BYOL\cite{BYOL}                                        & 1000                    & 78.4                           & 89.0                           & 54.0                           & 86.6                           \\
SwAV\cite{SwAV}                                        & 1000                    & 78.5                           & 89.9                           & -                              & -                              \\
Barlow Twins\cite{Barlow}                                & 1000                    & 79.2                           & 89.3                           & 54.1                           & 86.2                           \\
VICReg\cite{vicreg}                                      & 1000                    & 79.4                           & 89.5                           & 54.3                           & 86.6                           \\ 
SiamMM (ours)                                 & \textbf{200}         & \textbf{79.4}  & \textbf{89.2}  & \textbf{53.2}  & \textbf{87.3} \\
\bottomrule
\end{tabular}
\end{sc}
\end{small}
\end{center}
\end{table}

\minisection{Transfer Learning}
Following the setup of \cite{PIRL}, we train a linear classifier on top of the frozen SiamMM backbone on other downstream tasks: Places205 \cite{zhou2014learning} scene classification and VOC07 \cite{everingham2009pascal} multi-label image classification. For Places205 dataset, we train a fully-connected layer followed by softmax and report the top-1 accuracy (in \%); for VOC07, we train a linear SVM and report mAP in Table~\ref{semi-sup}. Our SiamMM achieves results on par with the state-of-the-art on both downstream tasks.

\minisection{Clustering Evaluation}
We evaluate the clustering quality by computing the Adjusted Mutual Information (AMI) \cite{vinh2009information} between clustering index obtained by different methods and the ground truth label on ImageNet. 
In the second experiment, we map to the most frequent true class label to every cluster and evaluate the top-1 accuracy by following \cite{MoCLR}. 
Table~\ref{tab:cluster_ami} demonstrates the higher quality of the clusters generated by SiamMM that significantly exceeds the previous unsupervised learning methods.

\begin{table}[t]
\caption{\textbf{Evaluation of Quality of Clustering Results.} Following \cite{PCL}, Adjusted Mutual Information (AMI) are reported under 200 epochs pretraining on ImageNet with $25k$ clusters. Following \cite{MoCLR}, every clusters mapped to the most frequent true class label to compute top-1 accuracy (\%).}
\label{tab:cluster_ami}
\begin{center}
\begin{small}
\begin{sc}
\begin{tabular}{lcc}
\toprule
Method            & AMI ($K=25k$)   & Acc ($K=1k$)\\ \midrule
DeepCluster\cite{Deepcluster}    & 0.28  & - \\
MoCo\cite{MoCo}           & 0.29 & - \\
SwAV\cite{SwAV}           & 0.40 & - \\
PCL\cite{PCL}            & 0.41  & - \\
ProPos\cite{Propos}      & 0.53  & - \\ 
SimCLR\cite{SimCLR}      &  -    &  33.3 \\
BYOL\cite{BYOL}          &  -    &  51.0 \\
MoCLR\cite{MoCLR}             &  -    &  51.6 \\ 
SiamMM (ours)            & \textbf{0.60}  &  \textbf{56.3} \\ 
\bottomrule
\end{tabular}
\end{sc}
\end{small}
\end{center}
\end{table}

\section{Ablation Experiments}
\subsection{Instance-wise v.s Cluster-wise Losses}
To evaluate the importance of the clustering objective, we ablate the instance loss and only implement the loss (\ref{eq:soft}). The linear evaluation top-1 accuracy drops to 71. 8\% under 200-epoch pre-training, though it is still better than that of SimSiam (see row \#1 \& \#2 in Table~\ref{tab:instance}). 

\begin{table}[t]
\caption{Instance-wise and cluster-wise losses ablation (row 1-2);  effect of instance-wise contrastive loss of Moco v3~\cite{mocov3} (row 3-5).}
\label{tab:instance}
\begin{center}
\begin{small}
\begin{sc}
\begin{tabular}{lcccc}
\toprule
Method     & \multicolumn{1}{l}{inst.-wise} &\multicolumn{1}{l}{neg. inst.} & \multicolumn{1}{l}{clus.-wise} & \multicolumn{1}{l}{Top-1}  \\ \midrule
SimSiam & X                                 &                      &            & 70.0                                                  \\
SiamMM(w/o i)       &                                   &                  &   X           & 71.8                                   \\ \midrule
Moco v3      &  X     & X    &     & 71.0  \\
SiamMM(w/ n)       &  X     & X    &  X  & 73.1  \\
SiamMM       & X                                 &                   &   X           & 73.2                                      \\
\bottomrule
\end{tabular}
\end{sc}
\end{small}
\end{center}
\end{table}

\subsection{Instance-wise Negative Samples}
The proposed method can incorporate instance-wise self-supervised learning objectives. We replace the $L_{inst}$ in (\ref{eq:inst_loss}) with contrastive loss of MoCo v3~\cite{mocov3} and denote the new additive loss as ``SiamMM w/ negative inst'' (see row\#4 in Table~\ref{tab:instance}). It shows that the MoCo v3 can get boosted by more than 2\% after adding our cluster-wise loss (\ref{eq:soft}). 
Besides, as ``SiamMM w/ negative inst'' achieves similar result to SiamMM, it shows that negative instance is not needed.

\subsection{Cluster Merging}
We evaluated the performance and training time of the model with and without the cluster-merging strategy. Table~\ref{tab:merge} presents the top-1 accuracy of linear evaluation for various numbers of clusters. Initiating with a substantial number like $100k$ and progressively merging down to a smaller cluster count markedly enhances model performance compared to fixed-number clustering algorithms when the final number of clusters is similar. Furthermore, in comparison to employing a fixed large number of clusters, the training time is approximately halved without a significant loss in accuracy.  In comparison, SimSiam requires only 65\% of the training time of SiamMM (starting from 100k clusters), but it results in a 3.2\% drop in performance compared to SiamMM.
\begin{table}[t]
\caption{\textbf{Fix or merge the number of clusters during training.} Top-1 accuracy (\%) of linear evaluation on ImageNet, and training time compared with the baseline (row \#1) are reported.}
\label{tab:merge}
\begin{center}
\begin{small}
\begin{sc}
\begin{tabular}{llcc}
\toprule
\multicolumn{2}{l}{final \# of clusters} & TOP-1    & Train Time  \\ \midrule
\multirow{2}{*}{1k}    & fixed     & 71.6 & 1x     \\
                       & merged    & 73.0 &  1.15x \\ 
\multirow{2}{*}{5k}    & fixed     & 72.4 &  1.1x     \\
                       & merged    & 73.2 &  1.20x \\ 
100k                   & fixed     & 73.2 &  2.33x \\
\bottomrule
\end{tabular}
\end{sc}
\end{small}
\end{center}
\end{table}

\subsection{Initial Number of Clusters for Merging}
\begin{figure}[t]
\begin{center}
\centerline{\includegraphics[scale=0.45]{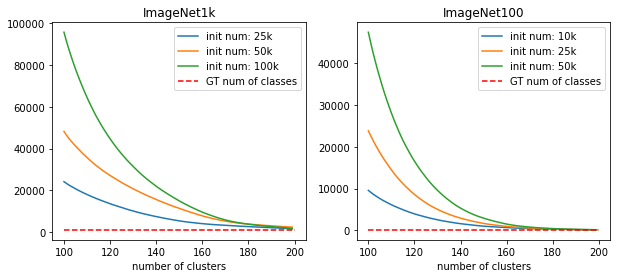}}
\caption{Starting from different initial numbers, the number of clusters converge almost to the true number of cluster in the datasets (left: ImageNet1k\cite{deng2009imagenet}; right: ImageNet100\cite{imagenet100}).}
\label{fig:merging_time}
\end{center}
\end{figure}

We conduct a comparative analysis of the performance of the model based on three different numbers of initial clusters, considering both hard and soft cluster assignments. Regardless of the initial number of clusters, the final number of clusters converges to a to almost the same number of classes included in ImageNet, as depicted in Figure~\ref{fig:merging_time}. We extended this analysis to two subsets of ImageNet—ImageNet100\cite{imagenet100} and TinyImageNet\cite{tinyimagenet}, which contain $100$ and $200$ classes, respectively—and found similar convergence patterns, underscoring the robustness of our approach across different datasets.

Table~\ref{tab:nnc} illustrates that initiating with a larger number of clusters and subsequently merging down increases the top-1 accuracy. This finding supports our hypothesis that overseeding clusters initially provides more opportunities to discover the ``correct'' centroids, resulting in enhanced clustering. However, increasing the cluster count from $50k$ to $100k$ yielded $0.1\%$ returns in accuracy, while largely increasing time complexity. Based on these findings, we recommend using $50k$ clusters as a balanced choice, offering substantial computational efficiency while maintaining strong performance. Additionally, we observed that the soft-assignment strategy improved top-1 accuracy by $0.2\%$ compared to the hard-assignment approach.

\begin{table}[t]
\caption{\textbf{Initial number of clusters.} Initiating with different number of clusters ($25k$, $50k$, $100k$) and merging down to a similar number of cluster count. Top-1 accuracy (\%) of linear evaluation on ImageNet.}
\label{tab:nnc}
\begin{center}
\begin{small}
\begin{sc}
\begin{tabular}{cccc}
\toprule
init \# of clusters & hard assign & soft assign  & train time \\ \midrule
25k                 & 72.6           & 72.8      & x   \\
50k                 & 72.9        & 73.1         & 1.1x\\
100k                & 73.0        & 73.2         & 1.3x\\
\bottomrule
\end{tabular}
\end{sc}
\end{small}
\end{center}
\end{table}

\begin{table}[t]
\caption{\textbf{Concentration Parameter and Consistent Update.} Top-1 accuracy (in \%) of linear evaluation on ImageNet under 200 epochs: w/ and w/o the per-cluster concentration parameter $\kappa$ (row \#1 \& \#3); w/ and w/o the consistent update (row \#2 \& \#3).}
\label{tab:non_uniform}
\begin{center}
\begin{small}
\begin{sc}
\begin{tabular}{lcccc}
\toprule
Method & Consist Update & Concent. $\kappa$ & TOP-1 \\ \midrule
SiamMM    & X      &    &  72.9 \\
SiamMM    &        & X  &  73.1   \\
SiamMM    & X      & X  &  73.2   \\
\bottomrule
\end{tabular}
\end{sc}
\end{small}
\end{center}
\end{table}

\begin{figure*}[h]
\centering
\begin{subfigure}{.5\textwidth}
  \centering
  \includegraphics[width=\linewidth]{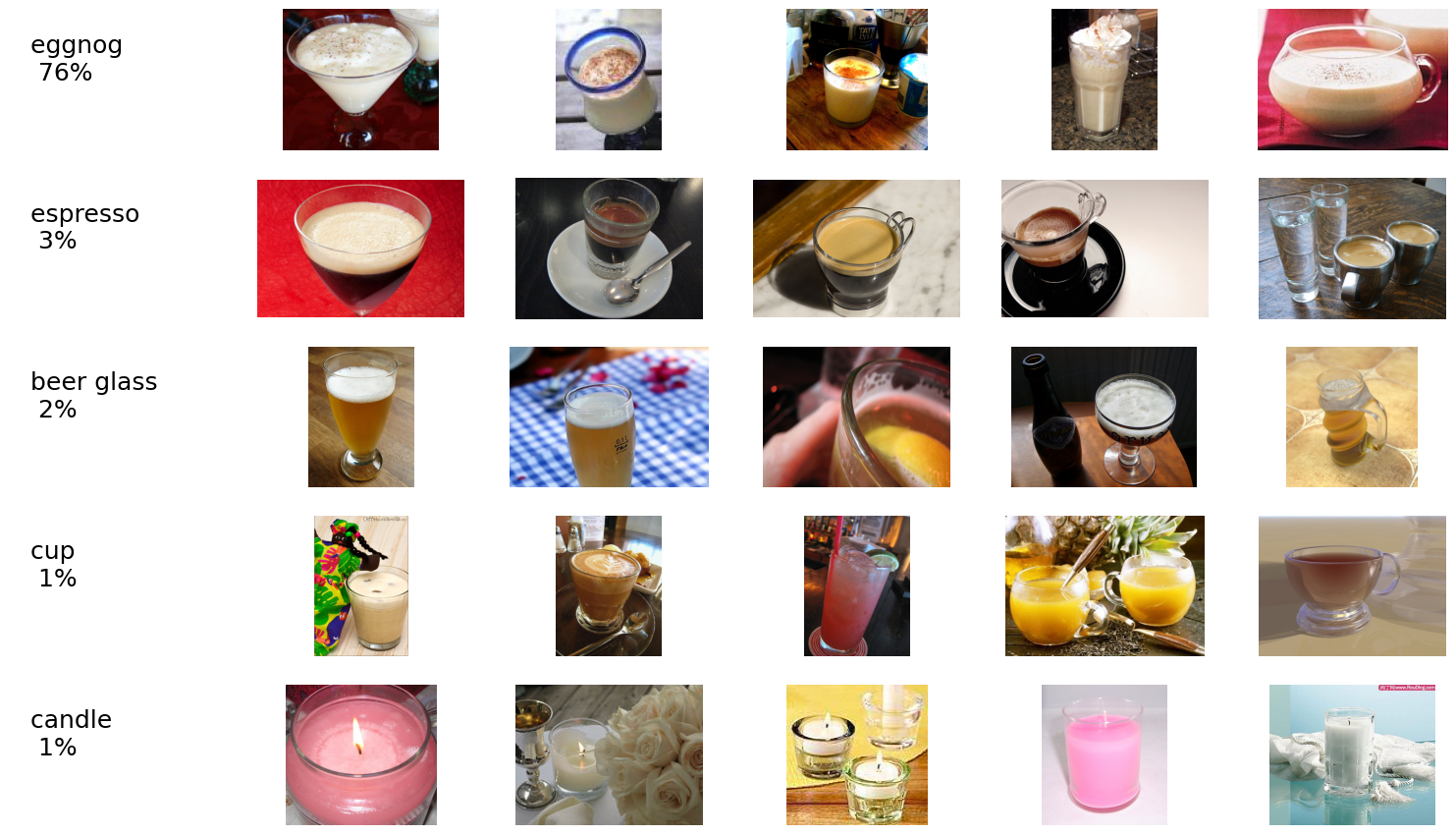}
\end{subfigure}%
\begin{subfigure}{.5\textwidth}
  \centering
  \includegraphics[width=\linewidth]{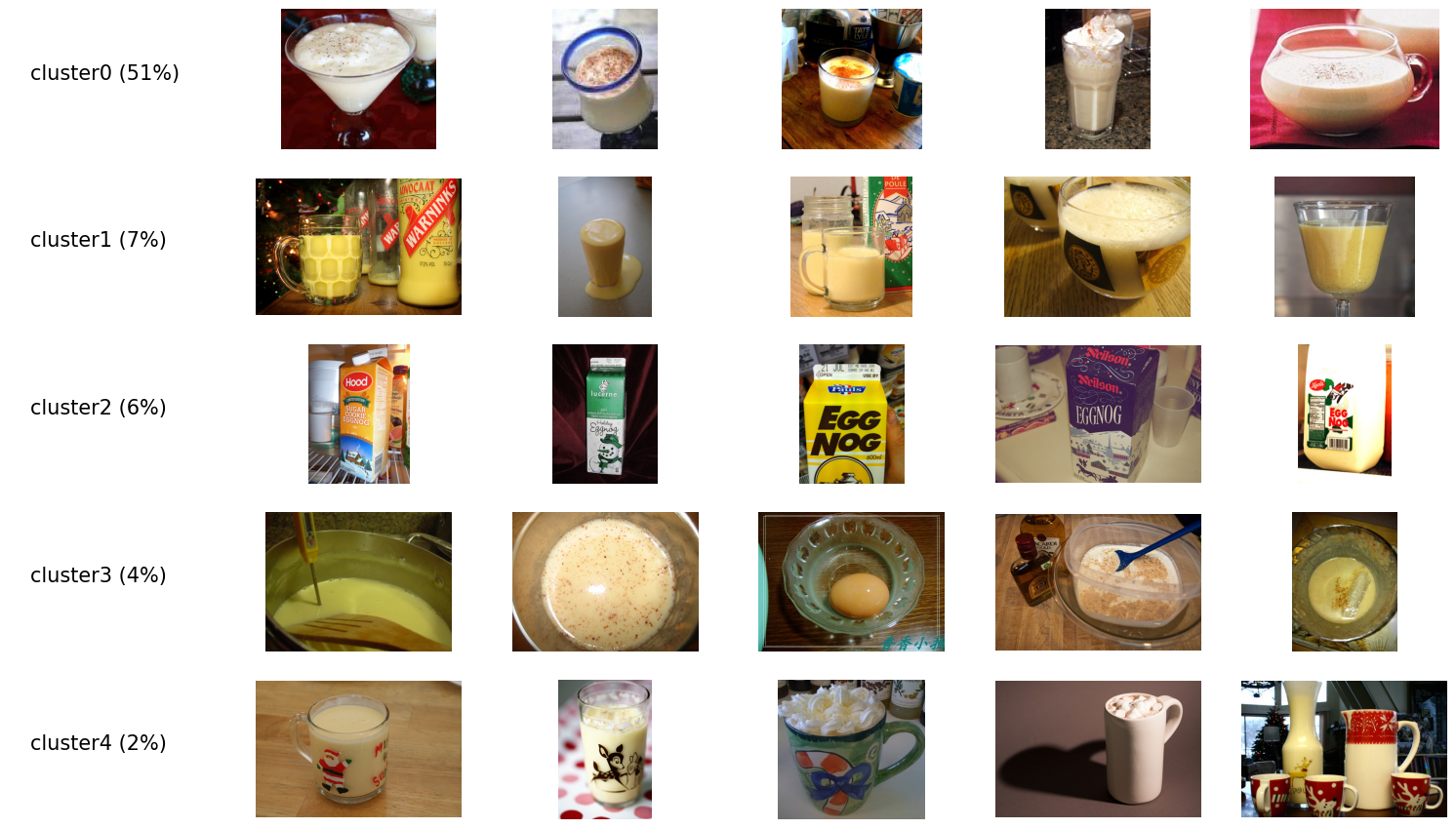}
\end{subfigure}
\caption{Visualization of Merged 1000 Clusters. Left: Images grouped under the same cluster label, where each row corresponds to a true class label in ImageNet. Right: Images grouped under class label ``eggnog'', where each row corresponds to a predicted clustering label.}
\label{fig.cluster_vis}
\end{figure*}

\subsection{Per-cluster Concentration Parameter}
We ablate the effect of having per cluster concentration parameter by replacing $\kappa_k$ with a constant number. Without taking into account the concentration scale per cluster, model performance drops by $0.3\%$ under 200-epoch pre-training (see Table~\ref{tab:non_uniform}).

\subsection{Consistent Centroid Updating}
As discussed in Section~\ref{sec:consistent}, we consistently update cluster centroids, propagating the learned centroids from the previous iteration to the next. In the ablation study, we adhere to the default setting \cite{PCL, Propos, Deepcluster}, randomly initializing cluster centroids when applying clustering per iteration.

Appendix~Figure \ref{fig:training_time} illustrates the training time for varying numbers of cluster re-initializations and with different fixed numbers of clusters. The training time for $200k$ clusters doubles when the number of re-initializations increases from $1$ to $5$. Furthermore, we observe a slight drop in model performance without applying the consistency strategy, as shown in Table~\ref{tab:non_uniform}. This highlights a major advantage of consistent update to inherit centroids from the previous epoch as initialization and avoiding re-initialization.

\subsection{Negative Sample is Unnecessary for SiamMM}

Many previous unsupervised learning methods borrow from contrastive methods and use large numbers of negative cluster centroids to form a contrastive or cross-entropy loss~\cite{PCL, Propos, Deepcluster, SwAV}. As discussed in Section~\ref{sec:neg_samp}, these negative centroids are not needed from a mixture model point of view. We thus reframe the cluster-wise loss (\ref{eq:soft}) (with $H=1$ for simplicity) by 
contrasting $v$ from its negative centroids as suggested by PCL \cite{PCL},
\begin{equation}
\label{eq:loss_nce1}
\Scale[0.95]{
  \mathcal{L}^{nce1} = - \log \frac{\exp(\hat{\kappa}_{\mathcal{A}_{+}} \hat{\mu}_{\mathcal{A}_{+}}^T v)}{\exp(\hat{\kappa}_{\mathcal{A}_{+}} \hat{\mu}_{\mathcal{A}_{+}}^T v) + \displaystyle\sum_{\mathcal{A}_{-}} \exp(\hat{\kappa}_{\mathcal{A}_{-}} \hat{\mu}_{\mathcal{A}_{-}}^T v)}
}
\end{equation}
\vspace{-0.mm}
where $\mathcal{A}_{+}$ and $\mathcal{A}_{-}$ indicates the positive centroid and negative centroids, respectively.

Alternatively, another sampling strategy is to sample negative image embedding $v^{-}$ that are out of the cluster $C_{A}$. So, the variant loss contrasts $v^+$ from the negative embeddings $v^{-}$ by
\begin{equation}
\label{eq:loss_nce2}
\begin{split}
\Scale[0.95]{
  \mathcal{L}^{nce2} = - \log \frac{\exp(\hat{\kappa}_{\mathcal{A}} \hat{\mu}_{\mathcal{A}}^T v^+)}{\exp(\hat{\kappa}_{\mathcal{A}} \hat{\mu}_{\mathcal{A}}^T v^+) + \displaystyle\sum_{v^-} \exp(\hat{\kappa}_{\mathcal{A}} \hat{\mu}_{\mathcal{A}}^T v^-)}
}
\end{split}
\end{equation}
\vspace{-0.mm}
where $v^{+}$ and $v^{-}$ are positive and negative image embedding, respectively. 
The evaluation results reported on Appendix~Table~\ref{tab:contrast} show that the two variations perform similarly to SiamMM. We conclude that simply integrating more negative samples does not help, possibly due to noise or redundant information.

\subsection{Cluster visualization}
We showcase visualizations of samples from the approximately $1000$ clusters learned by SiamMM, along with the corresponding ground truth labels for each image. Each cluster is characterized by the most common ground truth label within it in Figure~\ref{fig.cluster_vis}. In particular, despite the absence of truth labels during training, SiamMM learns clusters that exhibit a coherent alignment with ground truth labels. Furthermore, SiamMM reveals subtle texture or background differences that are consistently present across various image classes. These observations underscore the potential of SiamMM to improve data quality by identifying mislabeled images~\cite{northcutt2021pervasive} and highlighting more challenging negative samples.

\section{Conclusion}
We have presented a mixture model perspective on clustering in self- and unsupervised learning methods.
This framework allows us to develop a generalized framework and propose refined techniques for clustering in representation learning. 
Empirical evaluations confirm our individual insights and show that the combination of our parts result in learned representations performing better on a number of downstream tasks.
For future work, we would like to see how the mixture model perspective generalizes on classification datasets with class imbalance, detection, and video.

\clearpage
\newpage
\bibliographystyle{assets/plainnat}
\bibliography{paper}

\clearpage
\newpage
\beginappendix

\section{Algorithm Pseudocode}

\begin{algorithm}[h]
   \caption{SiamMM Pseudocode}
   \label{alg}
\begin{algorithmic}
    \State \# $\mathcal{E}$: backbone + projection MLP
    \State \# $\mathcal{E}_m$: momentum encoder of $\mathcal{E}$
    \State \# $g$: prediction MLP
   \State {\bfseries Input:} training data $X$, number of cluster $K$
   \State {\bfseries Initial:} cluster centroids $\{r_k\}^K_{k=1}$
   \While{not Max Epochs}
   \State {\bfseries E-step:} Update assignment $\mathcal{A}(.)$ by clustering embeddings on the centroid $\{r_k\}^K_{k=1}$ from the last iteration
   \State {\bfseries M-step:} Update $\mu_k$ and $\kappa_k$ for $k=1,2,...,K$
   \For{$x$ in $X$} 
   \State $x_1$, $x_2$ = $\mathrm{aug}(x)$, $\mathrm{aug}(x)$ \# image augmentation
   \State $v_1$, $v_2$ = $g(\mathcal{E}(x_1))$, $g(\mathcal{E}(x_2))$
   \State $v_1^m$, $v_2^m$ = $\mathcal{E}_m(x_1)$, $\mathcal{E}_m(x_2)$
   \State $\mathcal{L} (v_1,v_2,v_1^m,v_2^m | \mu_{\mathcal{A}(x)}, \kappa_{\mathcal{A}(x)})$ \# loss in (\ref{eq:tol_loss})
   \State update($\mathcal{E}$,$g$) \# optimization update
   \State update($\mathcal{E}_m$) \# momentum update
   \EndFor
   \State {\bfseries Merge:} Merge centroids $\{r_k\}^K_{k=1}$ based on (\ref{eq:merge})
   \State \quad\quad\quad\, Update number of clusters $K$
   \EndWhile
\end{algorithmic}
\end{algorithm}

\section{Relation to other methods}
We compare the proposed method with the previous state-of-the-art unsupervised learning method such as PCL~\cite{PCL}, SwAV~\cite{SwAV}, and ProPos~\cite{Propos}. 

\minisection{PCL}
Both PCL~\cite{PCL} and SiamMM attempt to improve the within-cluster compactness in an EM manner. However, PCL is based on an InfoNCE instance-wise contrastive loss requiring negative samples. As a result, PCL can suffer from the class collision issue \cite{Propos} from contrasting false negative instances. SiamMM does not require negative centroids (Section~\ref{sec:neg_samp}).

\minisection{SwAV} 
Though SwAV~\cite{SwAV} and SiamMM can be implemented by soft-assignment clustering, they represent different conceptual approaches to representation learning. SwAV poses clustering as a prediction problem, using the embedding of one augmented view to fit the assignment distribution of the other view, while SiamMM directly models the clustering distribution by maximizing the likelihood of a mixture model. 

\minisection{ProPos}
ProPos~\cite{Propos} and SiamMM have similarities in using non-contrastive losses as the instance-wise objective. The primary difference is that ProPos implements clustering on two augmented views to compute paired cluster centroids, with an infoNCE contrastive loss requiring a large number of negative centroids. However, SiamMM has a more straightforward interpretation of the mixture model and is more sample efficient in not requiring negative centroids.

\section{Additional Implementation Details}
\subsection{Details on parameter estimation}
\label{apx:pca}
The loss function of SiamMM attempts to pull image embeddings closer to the nearest cluster centroids on a sphere. The mean estimator $\hat{\mu}$ is a normalized cluster centroid. The concentration parameter $\kappa$ scales the gradient for each cluster, respectively. It tries to make the loosely distributed cluster more compact compared with the densely distributed clusters. Based on (\ref{eq:kappa}), the estimator includes the length of centroids vector $||r||$ and the dimensionality $d$. However, as the dimension of embedding might be correlated, it is non-trivial to estimate the dimensionality $d$.

One simple way to address this problem is to use PCA to reduce the dimension of $r$ and re-measure it length in a low-dimensional space. Based on our experiments, with only 150 principle components the $r^{PCA}$ can reserve more than $80 \%$ norm out of the original 256-dim $r$, so we rewrite (\ref{eq:kappa}) by 
\begin{equation}
\label{eq:kappa_pca}
\begin{split}
  \hat{\kappa_k}^{PCA} = \frac{||r_k^{PCA}||d^{PCA} - ||r_k^{PCA}||^3}{1-||r_k^{PCA}||^2}
 \end{split}
\end{equation}
where $d^{PCA}$ is the number of principle components selected to form $r_k^{PCA}$. Note that the dimension reduction has only been done in the estimation of $\kappa$. We still keep the original dimensionality of image representations. The discussion of dimension reduction in representation learning is out of the scope of this paper.

\section{Variants of SiamMM loss}
\label{sec:variant}

\begin{figure}[ht]
\label{fig:add_fig}
\begin{center}
\centerline{\includegraphics[scale=0.4]{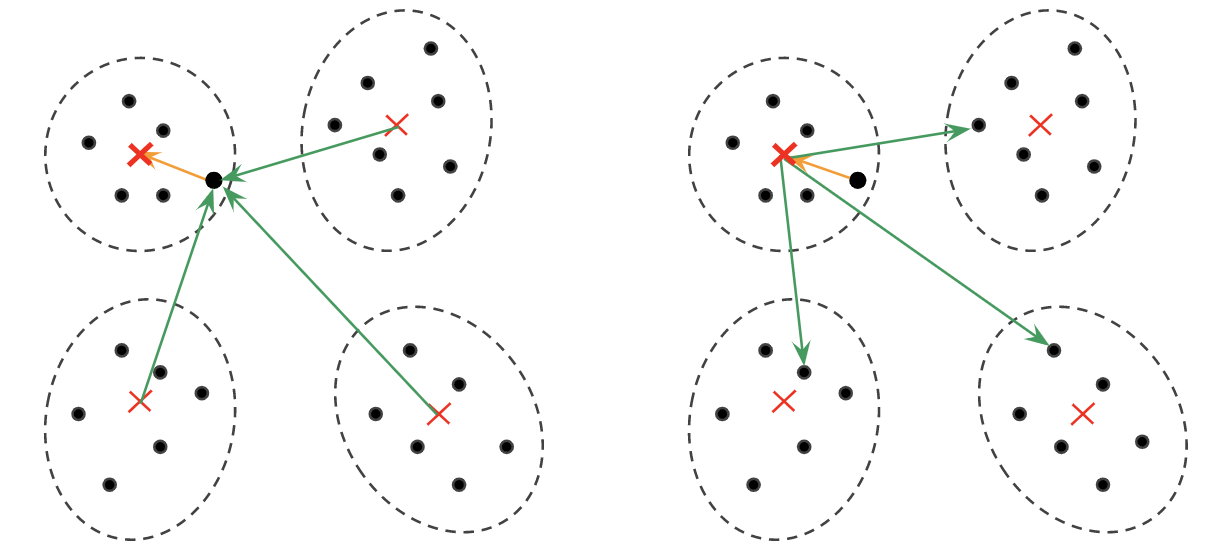}}
\caption{Illustration of the differenct negative sampling strategies; the dot and cross indicate the embedding of a data point and a cluster centroid, respectively; the blue arrow illustrates a positive pair, while the red arrow illustrates a negative pair. (Left) sampling negative cluster centroids (\ref{eq:loss_nce1}) either restored in a memory bank \cite{PCL} or from a batch size \cite{SwAV}; (Right) sampling negative data points out of a target cluster, introduced in (\ref{eq:loss_nce2})}.
\end{center}
\end{figure}

\begin{table}[h]
\caption{Linear evaluation on ImageNet for the variant losses of SiamMM; top-1 accuracy is reported (in \%); Compared with SiamMM, SiamMM+$\mathcal{L}^{nce1}$ (\ref{eq:loss_nce1}) introduces negative centroids whose number is chosen as $|\mathcal{A}_{-}|=16000$ as suggested in \cite{PCL}; SiamMM+$\mathcal{L}^{nce2}$ (\ref{eq:loss_nce2}) introduces additional negative instances.}
\label{tab:contrast}
\begin{center}
\begin{small}
\begin{sc}
\begin{tabular}{lccc}
\toprule
Method        & \multicolumn{1}{l}{Neg. inst.} & \multicolumn{1}{l}{Neg. cent.} & \multicolumn{1}{l}{Top-1} \\ \midrule
SiamMM $\mathcal{L}^{nce1}$                               &                                & X                              & 72.3                      \\
SiamMM $\mathcal{L}^{nce2}$                              & X                              &                                & 72.2                      \\
SiamMM                                      &                                &                                & 72.2         \\            
\bottomrule
\end{tabular}
\end{sc}
\end{small}
\end{center}
\end{table}

\newpage

\section{Additional Figures}

\begin{figure}[h]
\begin{center}
\centerline{\includegraphics[scale=0.4]{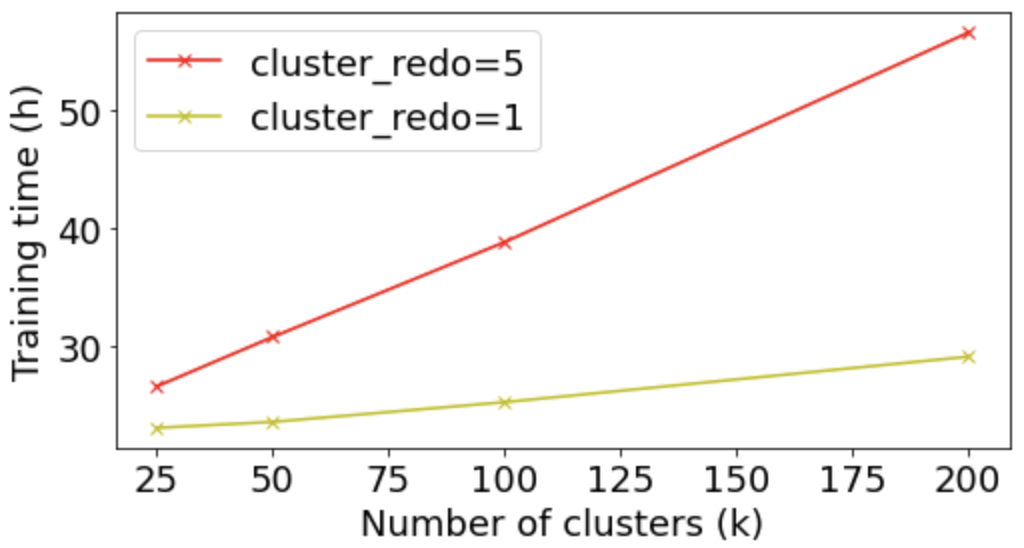}}
\caption{Training time per $100$ epochs for different numbers of cluster re-initializations and with different number of clusters.}
\label{fig:training_time}
\vspace{-0.45in}
\end{center}
\end{figure}

\begin{figure}[h]
\begin{center}
\centerline{\includegraphics[scale=0.4]{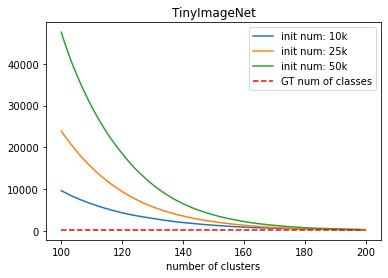}}
\caption{The number of clusters converge to almost the same number of classes included in TinyImageNet.}
\label{fig:training_time}
\vspace{-0.45in}
\end{center}
\end{figure}

\begin{figure}[h]
\begin{center}
\centerline{\includegraphics[scale=0.4]{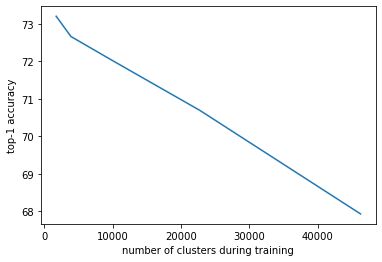}}
\caption{Read from right to left, the top-1 accracy increases linearly as the number of clusters converges on par with the true number of classes during the training process.}
\label{fig:training_time}
\vspace{-0.45in}
\end{center}
\end{figure}


\begin{figure*}[h]
\centering
\begin{subfigure}{.5\textwidth}
  \centering
  \includegraphics[width=\linewidth]{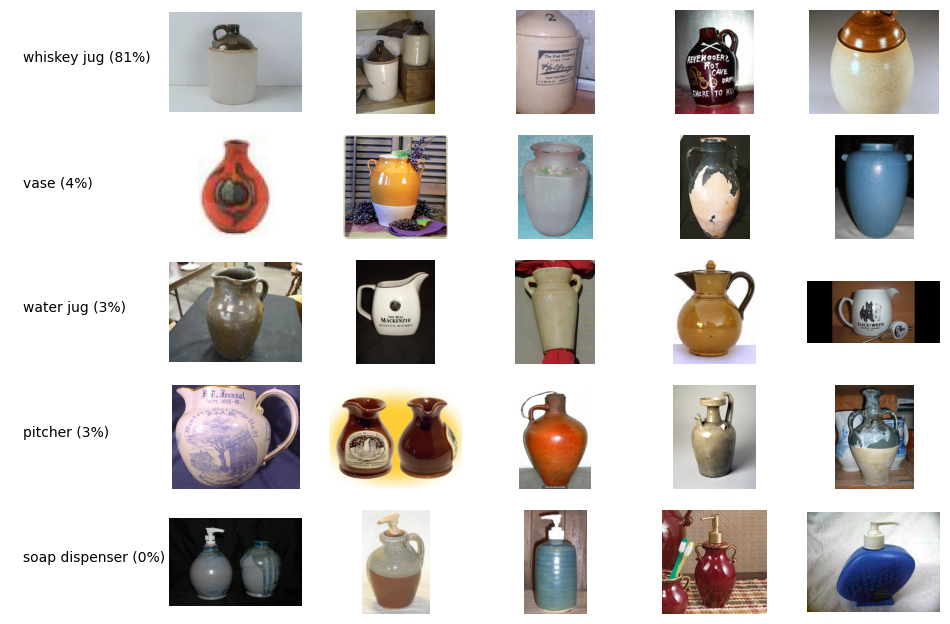}
\end{subfigure}%
\begin{subfigure}{.5\textwidth}
  \centering
  \includegraphics[width=\linewidth]{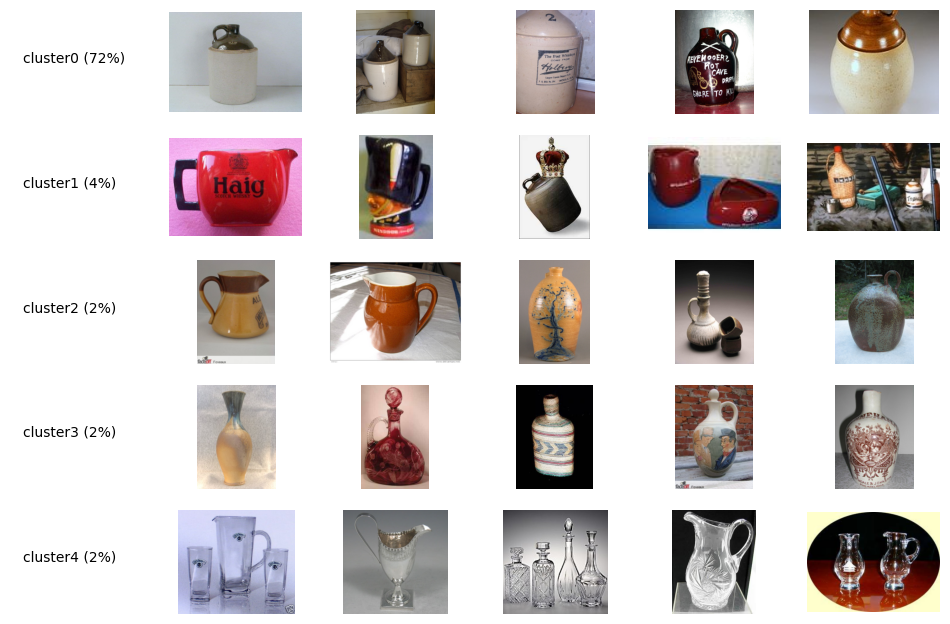}
\end{subfigure}
\caption{Visualization of Merged 1000 Clusters. Left: Images grouped under the same cluster label, where each row corresponds to a true class label in ImageNet. Right: Images grouped under class label ``whiskey jar'', where each row corresponds to a predicted clustering label.}
\end{figure*}

\begin{figure*}[h]
\centering
\begin{subfigure}{.5\textwidth}
  \centering
  \includegraphics[width=\linewidth]{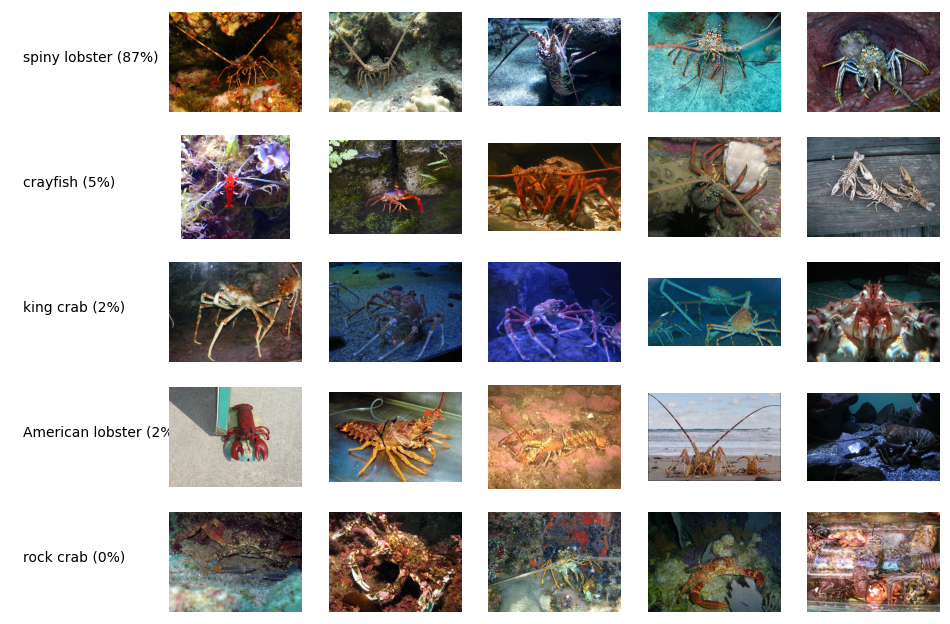}
\end{subfigure}%
\begin{subfigure}{.5\textwidth}
  \centering
  \includegraphics[width=\linewidth]{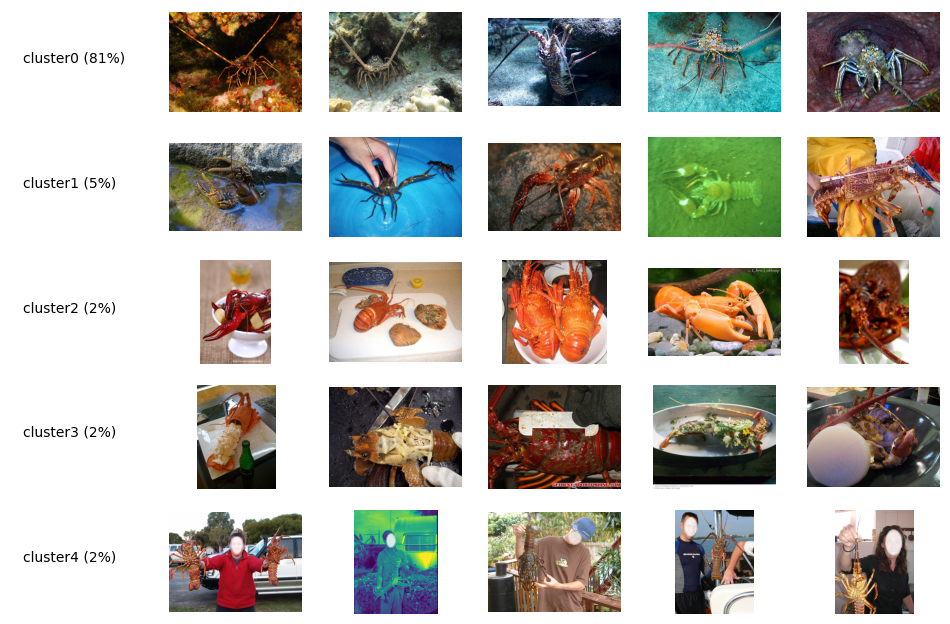}
\end{subfigure}
\caption{Visualization of Merged 1000 Clusters. Left: Images grouped under the same cluster label, where each row corresponds to a true class label in ImageNet. Right: Images grouped under class label ``spiny lobster'', where each row corresponds to a predicted clustering label. Identifiable faces are blurred in the images.}
\end{figure*}

\begin{figure*}[h]
\centering
\begin{subfigure}{.5\textwidth}
  \centering
  \includegraphics[width=\linewidth]{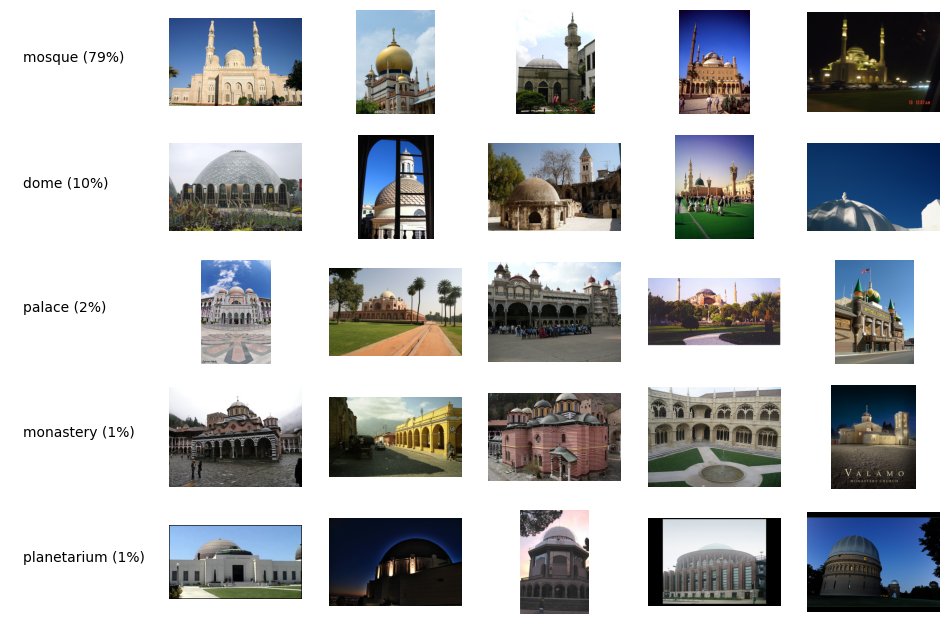}
\end{subfigure}%
\begin{subfigure}{.5\textwidth}
  \centering
  \includegraphics[width=\linewidth]{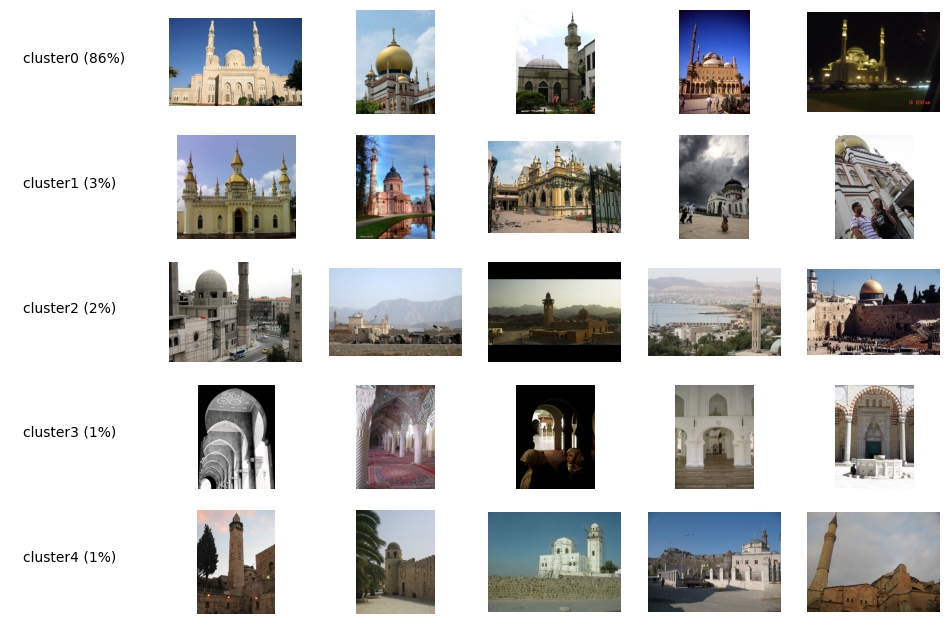}
\end{subfigure}
\caption{Visualization of Merged 1000 Clusters. Left: Images grouped under the same cluster label, where each row corresponds to a true class label in ImageNet. Right: Images grouped under class label ``mosque'', where each row corresponds to a predicted clustering label.}
\end{figure*}


\end{document}